%% file: main.tex
\documentclass[conference]{IEEEtran}
\IEEEoverridecommandlockouts
\def\BibTeX{{\rm B\kern-.05em{\sc i\kern-.025em b}\kern-.08em
    T\kern-.1667em\lower.7ex\hbox{E}\kern-.125emX}}
\usepackage{cite}
\usepackage{amsmath,amssymb,amsfonts,pifont}
\usepackage[normalem]{ulem}
\usepackage{enumitem}
\usepackage{algorithmic}
\usepackage{graphicx}
\usepackage{textcomp}
\usepackage{url}
\usepackage{xcolor}
\usepackage{multirow}
\usepackage{longtable}
\usepackage{afterpage}
\usepackage{rotating}
\usepackage{pdflscape}
\usepackage{booktabs}
\usepackage{makecell}
\usepackage{needspace}
\usepackage{array}
\usepackage{float}
\usepackage{subcaption}
\usepackage{colortbl}
\usepackage[acronym]{glossaries}
\usepackage[left=0.64in,right=0.64in,top=0.75in]{geometry}

\makeglossaries
\input{lib/acronyms}

\input{lib/defines}
\makeatletter
\newcommand{\linebreakand}{%
  \end{@IEEEauthorhalign}
  \hfill\mbox{}\par
  \mbox{}\hfill\begin{@IEEEauthorhalign}
}
\makeatother

\begin{document}

\title{Machine-Learning-Based Classification of Radio Frequency Building Loss\\ %*\\
}

\author{
    \IEEEauthorblockN{
        Jiayi Tan\IEEEauthorrefmark{1},  
        Neelabhro Roy\IEEEauthorrefmark{2},
        James Gross\IEEEauthorrefmark{2},
        Rohit Chandra\IEEEauthorrefmark{1} and 
        Tsao-Tsen Chen\IEEEauthorrefmark{1}
    } 
    \IEEEauthorblockA{
        \IEEEauthorrefmark{1}%\textit{Department of Radio Access Propagation} \\
        \textit{Ericsson AB,         Stockholm, Sweden }\\
        %Stockholm, Sweden \\
    }
    \IEEEauthorblockA{
        \IEEEauthorrefmark{2}%\textit{School of Electrical Engineering \& Computer Science} \\
        \textit{KTH Royal Institute of Technology,         Stockholm, Sweden }\\
        Email: \{jiayi.tan, rohit.chandra, jason.t.chen\}@ericsson.com, \{nroy, jamesgr\}@kth.se\\
    } 
}
\maketitle

\begin{abstract}
\input{chapters/abstract}
\end{abstract}


\section{Introduction}
\label{sec:intro}
\input{chapters/introduction}

%\section{Related Work}
%\input{chapters/related_work}
%\label{sec:related_work}

\section{System Model and Problem Statement}
\label{sec:system_model}
\input{chapters/system_model}

\section{Proposed Methodology}
\label{sec:proposed_method}
\input{chapters/methods}

\section{Results and Discussions}
\label{sec:results}
\input{chapters/results}

\section{Conclusions}
\label{sec:conclusion}
\input{chapters/conclusions}

%\section*{Acknowledgment}
%
%The preferred spelling of the word ``acknowledgment'' in America is without 
%an ``e'' after the ``g''. Avoid the stilted expression ``one of us (R. B. 
%G.) thanks $\ldots$''. Instead, try ``R. B. G. thanks$\ldots$''. Put sponsor 
%acknowledgments in the unnumbered footnote on the first page.

%\section*{References}
%\printbibliography[heading=bibintoc]
%\newpage
\bibliographystyle{ieeetr}
\bibliography{references}
\clearpage

% Appendix ===============================
%\onecolumn
%\appendix
%\input{chapters/appendix}

\end{document}

%% file: lib/acronyms.tex
\newacronym{I2I}{I2I}{Indoor-to-Indoor}
\newacronym{O2I}{O2I}{Outdoor-to-Indoor}
\newacronym{Wi-Fi}{Wi-Fi}{Wireless Fidelity}
\newacronym{Low-E}{Low-E}{Low-Emissivity}
\newacronym{IoT}{IoT}{Internet of Things}
\newacronym{5G}{5G}{5th Generation}
\newacronym{3D}{3D}{Three-Dimensional}
\newacronym{OSM}{OSM}{OpenStreetMap}
\newacronym{ML}{ML}{Machine Learning}
\newacronym{PL}{PL}{Path Loss}
\newacronym{Tx}{Tx}{Transmitter}
\newacronym{Rx}{Rx}{Receiver}
\newacronym{RanF}{RanF}{Random Forest}
\newacronym{XGBoost}{XGBoost}{Extreme Gradient Boosting}
\newacronym{LightGBM}{LightGBM or LGBM}{Light Gradient Boosting Machine}
\newacronym{BEL}{BEL}{Building Entry Loss}
\newacronym{RF}{RF}{Radio Frequency}
\newacronym{UE}{UE}{User Equipment}
\newacronym{GPS}{GPS}{Global Positioning System}
\newacronym{EARFCN}{EARFCN}{E-UTRA Absolute Radio Frequency Channel Number}

\newacronym{AoA}{AoA}{Angle of Arrival}
\newacronym{FSPL}{FSPL}{Free Space Path Loss}
\newacronym{RSRP}{RSRP}{Reference Signal Received Power}
\newacronym{SINR}{SINR}{Signal-to-Interference-plus-Noise Ratio}
\newacronym{RSRQ}{RSRQ}{Reference Signal Received Quality}
\newacronym{RSSI}{RSSI}{Received Signal Strength Indicator}
\newacronym{MIMO}{MIMO}{Multiple-Input Multiple-Output}
\newacronym{mmWave}{mmWave}{millimeter-Wave}
\newacronym{KPI}{KPI}{Key Performance Indicator}
\newacronym{O-RAN}{O-RAN}{Open Radio Access Network}
\newacronym{2D}{2D}{Two-Dimensional}
\newacronym{2G}{2G}{2th Generation}
\newacronym{3G}{3G}{3th Generation}
\newacronym{4G}{4G}{4th Generation}
\newacronym{6G}{6G}{6th Generation}
\newacronym{RRM}{RRM}{Radio Resource Management}
\newacronym{CAPEX}{CAPEX}{Capital Expenditure}
\newacronym{OPEX}{OPEX}{Operational Expenditure}
\newacronym{QoS}{QoS}{Quality of Service}
\newacronym{QoC}{QoC}{Quality of Control}
\newacronym{QoE}{QoE}{Quality of Experience}
\newacronym{COST231}{COST231}{COST 231 Multi-Wall Model}
\newacronym{DT}{DT}{Digital Twin}
\newacronym{PM}{PM}{Performance Management}
\newacronym{MDT}{MDT}{Minimization of Drive Test}
\newacronym{LOS}{LOS}{Line-of-Sight}
\newacronym{NLOS}{NLOS}{Non-Line-of-Sight}
\newacronym{BS}{BS}{Base Station}
\newacronym{ISP}{ISP}{Internet Service Provider}
\newacronym{LLM}{LLM}{Large Language Model}
\newacronym{LiDAR}{LiDAR}{Light Detection and Ranging}
\newacronym{RT}{RT}{Ray Tracing}
\newacronym{GBSM}{GBSM}{Geometry-Based Stochastic Model}
\newacronym{LDPL}{LDPL}{Log-Distance Path Loss}
\newacronym{3GPP}{3GPP}{3rd Generation Partnership Project}
\newacronym{NR}{NR}{New Radio}
\newacronym{LTE}{LTE}{Long Term Evolution}
\newacronym{FI}{FI}{Floating-Intercept}
\newacronym{FAF}{FAF}{Floor Attenuation Factor}
\newacronym{EPC}{EPC}{Energy Performance Certificate}
\newacronym{SAP}{SAP}{Standard Assessment Procedure}
\newacronym{RdSAP}{RdSAP}{Reduced data Standard Assessment Procedure}
\newacronym{LBSM2}{LBSM2}{London Building Stock Model 2}
\newacronym{MICE}{MICE}{Multiple Imputation by Chained Equations}
\newacronym{SL}{SL}{Supervised Machine Learning}
\newacronym{SSL}{SSL or semi-SL}{Semi-Supervised Machine Learning}
\newacronym{PID}{PID}{Personally Identifiable Data}
\newacronym{SDK}{SDK}{Software Development Kit}
\newacronym{MMPP}{MMPP}{Mean Maximum Predicted Class Probability}
\newacronym{ODbL}{ODbL}{Open Database License}
\newacronym{OGL v3.0}{OGL v3.0}{Open Government Licence v3.0}

%% file: lib/defines.tex
\newcommand{\eg}{e.g.}

\newcommand{\etal}{et~al.}

%% file: chapters/abstract.tex
Accurate modeling of outdoor-to-indoor (O2I) and indoor-to-indoor (I2I) signal loss is important for improving indoor wireless network performance in dense urban areas. Traditional on-site measurements are expensive, time-consuming, and difficult to conduct across wide regions. Real-world datasets also tend to be noisy and imbalanced, which makes signal loss prediction challenging. This study presents a machine learning framework for classifying radio frequency (RF) building loss. The framework combines passively collected, crowdsourced user equipment (UE) data from 3GPP-compliant networks with public building information. We evaluated Random Forest, XGBoost, LightGBM, and a voting classifier using both supervised (SL) and semi-supervised learning (SSL). Compared to SL-only inference, the proposed SL and SSL framework improved both prediction accuracy and confidence under identical data constraints, achieving up to 12.6\% relative accuracy gain for O2I loss and 3.4\% for I2I loss, while reducing prediction entropy by up to 8.4\%.
Among the evaluated models, SSL XGBoost provided the most confident O2I loss classification, whereas SSL LightGBM achieved the best performance for I2I loss. These results demonstrate that the proposed approach provides a practical, data-driven alternative to traditional models, with promising potential to support better network planning and indoor coverage optimization.

%The resulting classifications were analyzed alongside RF and building features to assess robustness and consistency across different frequency bands and building types. All models performed well, with the semi-supervised XGBoost model showing the best for O2I classification and semi-supervised LightGBM outperforming others for I2I classification. 

%% file: chapters/keywords.tex
Building Loss, Outdoor-to-Indoor (O2I), Indoor-to-Indoor (I2I), 3GPP, Machine Learning, Semi-Supervised Learning, Pathloss model

%% file: chapters/introduction.tex
Rapid urbanization and the exponential growth of mobile data traffic have established ubiquitous wireless connectivity as a prerequisite for modern smart cities. However, reliable \gls{O2I} and \gls{I2I} coverage is increasingly compromised by energy-efficient building materials, as modern buildings adopt advanced insulation and \gls{Low-E} glazing, which significantly increases signal attenuation \cite{rudd14}.
At the same time, wireless services including \gls{IoT}, \gls{Wi-Fi}, and \gls{5G} cellular technology require robust indoor connectivity, conflicting with the physical barriers these materials introduce \cite{zhihao}. Meeting the high data-rate demands of these services drives deployment toward higher frequency bands, which experience greater path loss and are more strongly attenuated by modern building materials. Importantly, this attenuation is not only large in magnitude but also highly variable across different building types and construction materials, making it difficult to address through simple network densification. Adding more base stations without accounting for building-specific penetration loss can lead to inefficient spectrum use, increased interference, and persistent coverage gaps. These challenges motivate the need for scalable and accurate propagation models that capture building-specific attenuation effects.
%Propagation models serve this critical function by mathematically estimating path loss and signal strength as electromagnetic waves interact with the environment. By characterizing how signals traverse free space and penetrate obstacles, these models enable network planners to optimize coverage and capacity without solely relying on costly physical site surveys.
Existing propagation models can be broadly categorized into measurement-based, classical, and machine learning approaches. %\textcolor{red}{The following subsections review these approaches with a focus on their applicability to building-level O2I and I2I loss modeling in large-scale urban environments.}

\subsection{Measurement-Based Studies}
\label{subsec:measurement-based}

Measurement campaigns provide the empirical foundation for understanding O2I penetration. Studies have shown that modern materials like foil-backed insulation and Low-E glazing markedly reduce signal strength \cite{rudd14, owens22}. Specifically for O2I loss, Kodr \etal \cite{kodr24} demonstrated that standard models (\eg, 3GPP TR~38.901) often underestimate penetration loss in reinforced concrete structures. However, while these studies offer high fidelity, they are inherently site-specific and do not scale to city-wide tasks. 

\subsection{Deterministic and Empirical Models}
\label{subsec:deter-empi-model}

To achieve broader applicability, network planners often look toward deterministic or empirical models. Deterministic ray-tracing offers high accuracy for I2I and O2I loss by modeling 3D geometry \cite{ullah20, fuschini15, vitucci15, zhihao}, but its computational cost and reliance on high-fidelity indoor maps make it impractical for large-scale urban analysis. Conversely, empirical models like ITU-R P.1238 \cite{iturp1238} or COST231 \cite{cost231} are computationally efficient but are typically link-specific and require precise parameters (\eg, incident angle, wall thickness) that are rarely available in crowdsourced or large-scale urban datasets.

\subsection{Machine Learning (ML) Approaches}
\label{subsec:ml-model}

Recent \gls{ML} approaches, such as \gls{RanF} and \gls{XGBoost}, have emerged as scalable alternatives, often outperforming classical formulas by reducing prediction error and improving robustness in complex indoor and urban environments where diffraction, reflection, and clutter are difficult to model explicitly \cite{aldo23, mora20}. Despite their promise, existing ML models for O2I propagation struggle with generalization. Most are trained on features derived from specific simulation environments or controlled measurement setups, making them sensitive to the local topology of the training site. Furthermore, they frequently omit building-level structural proxies, such as construction age or footprint geometry, which are critical determinants of O2I attenuation in heterogeneous urban environments \cite{aldo23}.

\subsection{Contribution}
\label{subsec:contribution}

%In summary, prior work reveals three key limitations: 
%
%\begin{itemize}
%    \item \textbf{Measurement Studies:} Provide high accuracy but are site-specific, costly, and lack the throughput required for city-wide analysis.
%    \item \textbf{Deterministic/Empirical Models:} Rely on high-fidelity building data or rigid link-specific parameters, making them difficult to apply to heterogeneous urban environments where detailed material data is missing.
%    \item \textbf{Existing ML Approaches:} Often rely on site-specific geometric features, which limits their potential for transferability to diverse urban areas and fails to account for construction-driven attenuation.
%\end{itemize}

%To address these gaps, we propose a novel ML framework in this work, that bridges the gap between site-specific accuracy and city-wide applicability. By leveraging publicly available building metadata as structural and material proxies (\eg, building age, footprint geometry, and height). %, our approach moves beyond simple distance-based metrics. 

Table~\ref{tab:baseline_comparison} compares the proposed framework with representative O2I/I2I loss modeling approaches. Existing methods typically satisfy only a subset of desirable properties: measurement-based and ray-tracing approaches are material-aware but not scalable; empirical models are computationally efficient but coarse and not building-specific; and supervised ML approaches require dense labeled data and are generally not material-aware.
This work introduces a scalable, building-level O2I/I2I loss inference framework that is material-aware without requiring explicit material measurements or detailed indoor layouts. The proposed method combines (i) unsupervised loss quantization from passive UE measurements, (ii) semi-supervised learning to enable inference in data-sparse regions, and (iii) integration of public building metadata, \eg, building age, footprint geometry, and height, to capture structural effects at city scale. This combination enables practical large-scale indoor loss classification for urban wireless network planning.

\begin{table*}[t]
\centering
\normalsize
\renewcommand{\arraystretch}{1.25}
%\captionsetup{width=\columnwidth, justification=raggedright, singlelinecheck=false}
\caption{Comparison of the Proposed Framework with Representative Baseline Approaches for O2I/I2I Loss Modeling}
\label{tab:baseline_comparison}

\begin{tabular}{
>{\centering\arraybackslash}m{0.16\textwidth} |
>{\centering\arraybackslash}m{0.07\textwidth}
>{\centering\arraybackslash}m{0.07\textwidth}
>{\centering\arraybackslash}m{0.09\textwidth}
>{\centering\arraybackslash}m{0.07\textwidth}
>{\centering\arraybackslash}m{0.11\textwidth} |
>{\centering\arraybackslash}m{0.14\textwidth}
}
\hline
\textbf{Approach} &
\textbf{O2I/I2I} &
\textbf{Building-Level} &
\textbf{Material-Aware} &
\textbf{Scalable} &
\textbf{Low Comp. Cost} &
\textbf{Representative References} \\
\hline

Measurement-based studies
& \checkmark & \ding{55} & \checkmark & \ding{55} & \ding{55}
& \cite{kodr24} \\

\rowcolor[gray]{0.93}
Deterministic ray tracing
& \checkmark & \ding{55} & \checkmark & \ding{55} & \ding{55}
& \cite{fuschini15,vitucci15} \\

Empirical penetration models
& \checkmark & \ding{55} & \checkmark & \ding{55} & \checkmark
& \cite{iturp1238,cost231} \\

\rowcolor[gray]{0.93}
Supervised ML
& \checkmark & \ding{55} & \ding{55} & \checkmark & \checkmark
& \cite{aldo23,mora20} \\

\textbf{Proposed framework}
& \checkmark & \checkmark & \checkmark & \checkmark & \checkmark
& This work \\

\hline
\end{tabular}
\end{table*}

%% file: chapters/system_model.tex
This section defines the spatial environment, measurement models, and the mathematical representation of building-level O2I and I2I signal loss. These definitions provide the foundation for the methodology described in Section~\ref{sec:proposed_method}.

\subsection{Spatial Environment and Building Representation}
\label{subsec:sys_building}

We consider an urban area $\mathcal{A} \subset \mathbb{R}^2$ containing a finite set of buildings $\mathcal{B} = \{ b_1, b_2, \dots, b_{N_b} \}$.
%\begin{equation}
%    \mathcal{B} = \{ b_1, b_2, \dots, b_{N_b} \}.
%\end{equation}
Each building $b \in \mathcal{B}$ is represented by a polygonal footprint $\mathcal{P}_b \subset \mathcal{A}$ and an associated attribute vector $\mathbf{x}_b \in \mathbb{R}^d$, 
%\begin{equation}
%    \mathbf{x}_b \in \mathbb{R}^d,
%\end{equation}
which encodes structural, geometric, and semantic properties such as height, floor count, construction period, usage type, and energy-related indicators. 
Building attributes are obtained by fusing publicly available geospatial datasets, including \gls{OSM} \cite{osm_data} and \gls{LBSM2} \cite{lbsm_v2}. The datasets are used in compliance with their respective licenses, namely \gls{ODbL} for \gls{OSM} \cite{osm_license} and \gls{OGL v3.0} for \gls{LBSM2} \cite{ogl_v3}. Owing to heterogeneous data availability and reporting standards, the resulting feature vectors may contain missing or inconsistent entries, motivating the use of robust preprocessing and learning strategies.

\subsection{Crowdsourced UE Measurement Model}
\label{subsec:sys_measurement}

Let $\mathcal{S} = \{ s_1, s_2, \dots, s_{N_s} \}$ denote a set of passively collected \gls{UE} measurement samples. Each sample $s \in \mathcal{S}$ is associated with a reported horizontal position $\mathbf{p}_s \in \mathbb{R}^2$, a location accuracy radius $\sigma_s$, a serving cell identifier $c_s$, a carrier frequency band $f_s$, and a received signal strength measurement $P_s$ expressed as \gls{RSRP} [dBm]. Although vertical coordinates may be available, they are often device-dependent and exhibit significant uncertainty in crowdsourced datasets; therefore, only horizontal positioning is considered.

Due to the passive and device-dependent nature of the data, the true UE location is uncertain. Following common practice in crowdsourced positioning, this uncertainty is modeled as a bivariate Gaussian random variable $ \mathbf{P}_s \sim \mathcal{N}(\mathbf{p}_s, \sigma_s^2 \mathbf{I})$,
%\begin{equation}
%    \mathbf{P}_s \sim \mathcal{N}(\mathbf{p}_s, \sigma_s^2 \mathbf{I}),
%\end{equation}
where $\mathbf{I}$ denotes the $2 \times 2$ identity matrix. This is motivated by the fact that horizontal positioning errors in urban environments typically follow a Gaussian distribution where $x$ and $y$ error components are assumed independent and identically distributed (i.i.d.) based on the reported accuracy radius.

\subsection{Relative Signal Loss Formulation}
\label{subsec:sys_loss}

Absolute path loss cannot be directly estimated from crowdsourced UE measurements, as \gls{Tx}-side parameters (\eg, transmit power, antenna patterns, and exact base station locations) are not available. Consequently, this work focuses on \emph{relative signal loss}, derived from differences in received power between spatially proximate samples associated with the same serving cell and frequency band.

Consider two UE samples $s_1$ and $s_2$ served by the same cell $c$ at frequency $f$, with received powers $P_{s_1}$ and $P_{s_2}$. The relative signal loss between the two samples is defined as
\begin{equation}
\label{eq:rel_rf_loss}
    \ell_{s_1,s_2} = \frac{P_{s_1} - P_{s_2}}{d_{s_1,s_2}},
\end{equation}
where $d_{s_1,s_2}$ denotes the \gls{3D} Euclidean distance between the two UE locations. The normalization by distance compensates for geometric separation and isolates attenuation effects attributable to the propagation environment.

Relative \gls{O2I} losses are obtained from outdoor--indoor sample pairs, while relative \gls{I2I} losses are obtained from indoor--indoor pairs within the same building. These sample-level loss estimates are noisy and incomplete but provide indirect observations of building-level attenuation characteristics.

\subsection{Problem Statement}
\label{subsec:sys_problem}

Given a set of buildings $\mathcal{B}$ with attribute vectors $\{\mathbf{x}_b\}$ and a collection of crowdsourced UE measurements $\mathcal{S}$ with uncertain indoor--outdoor status, the objective is to infer building-level \gls{O2I} and \gls{I2I} attenuation characteristics. Rather than estimating absolute penetration loss values, which are poorly defined under partial observability, we formulate the task as a categorical inference problem. For each building $b \in \mathcal{B}$ and frequency band $f$, the goal is to estimate a loss label $y_{b,f} \in \mathcal{Y}$,
%\begin{equation}
%    y_{b,f} \in \mathcal{Y},
%\end{equation}
where $\mathcal{Y}$ denotes a finite set of attenuation categories.

In this work, $\mathcal{Y}$ is instantiated as a three-level quantization ($k=3$) corresponding to low, medium, and high relative loss. An unsupervised clustering approach was adopted to avoid manually defined thresholds, which are typically environment-dependent and difficult to transfer across regions. This choice enables scalable labeling in previously unobserved areas without requiring site-specific calibration or controlled measurements.

%We evaluated $k$-means clustering for $k \in [2,10]$ using the silhouette score. Across this range, the scores remained stable at approximately 0.50. For $k=3$, the silhouette score was about 0.51. Larger values of $k$ did not result in improved cluster separability and increased computational complexity without clear benefit. A two-class formulation ($k=2$) was examined in an internal pilot study and found to be overly coarse, as buildings with moderate and severe attenuation were grouped into a single class. Therefore, $k=3$ was selected as a practical compromise between resolution and scalability. 

%% file: chapters/methods.tex
\begin{figure*}[!t]
  \centering
  \includegraphics[width=\textwidth]{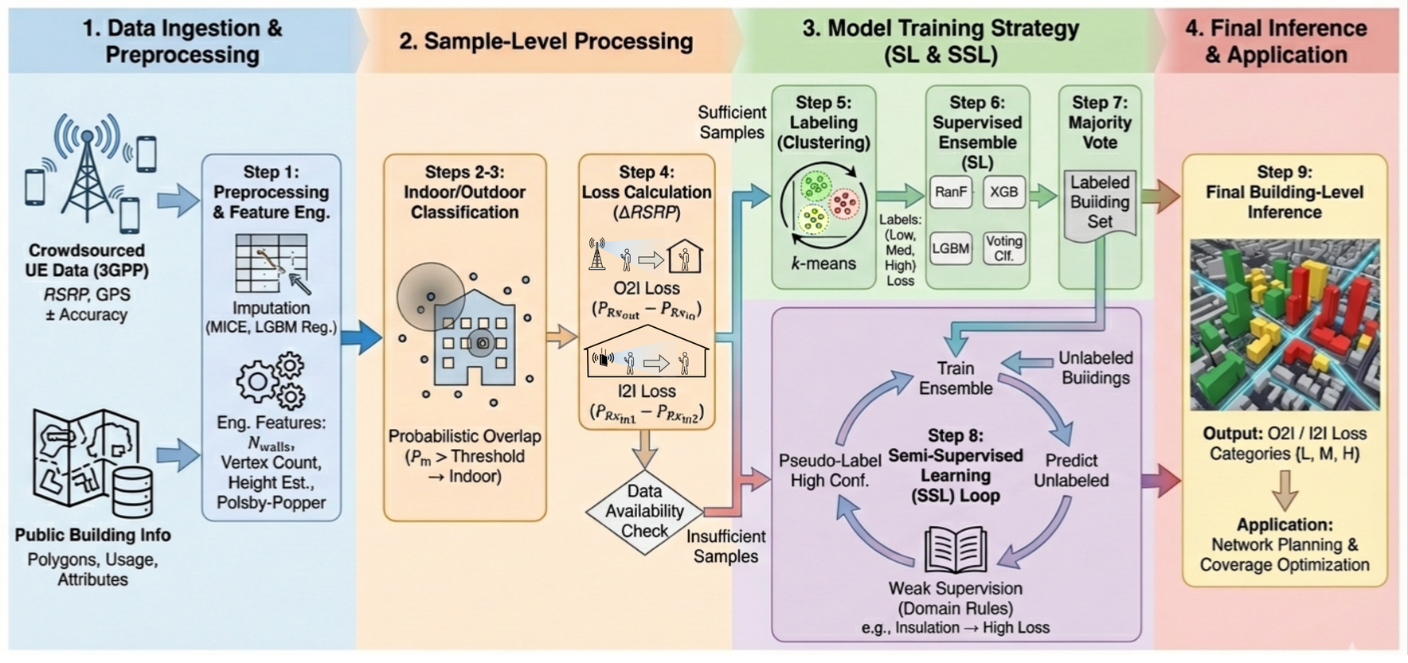}
  \caption{Proposed ML Framework for Building Loss Classification}
  \label{fig:method_workflow}
\end{figure*}

This section presents the proposed methodology for building-level \gls{O2I} and \gls{I2I} loss inference. An overview of the complete workflow is shown in Fig.~\ref{fig:method_workflow}. The methodology is organized into four stages: data sources and ingestion, preprocessing, loss construction and labeling, and model learning and inference.

%\vspace{-4mm}

\subsection{Data Sources and Ingestion}

The proposed framework was evaluated using crowdsourced UE measurements provided by Ookla/CellRebel \cite{ookla, cellrebel_portal}, collected over a three-month period (September - December 2022) within the City of London. The study area $\mathcal{A}$ covered approximately $5.9\,\text{km}^2$ and comprised $N_b = 5{,}366$ buildings with heterogeneous construction types, usage categories, and energy-performance characteristics. The used data sources were:
i) Crowdsourced UE measurements comprising approximately $7.6\times10^5$ samples, providing \gls{RSRP}, \gls{EARFCN}, and location metadata. ii) Public building footprint data from \gls{OSM}, defining building geometry. iii) Building metadata from \gls{LBSM2}, including construction period, usage type, and energy-performance indicators like \gls{EPC} ratings.

As mentioned in Section~\ref{subsec:sys_building}, these datasets were spatially aligned and fused at the building level using \gls{OSM} and \gls{LBSM2}, forming the basis for subsequent preprocessing and modeling steps.

\subsection{Data Preprocessing and Feature Engineering}

Missing or inconsistent building attributes were addressed using a combination of statistical imputation and multivariate regression. Features with limited missingness were imputed using simple summary statistics, while features with substantial missingness were handled using the \gls{MICE} framework with a \gls{LightGBM} regressor to capture non-linear dependencies. Rule-based constraints were applied to ensure physically plausible values. %(\eg, valid ranges for floors or heights).

To reduce dimensionality and improve interpretability, building usage types were grouped into four categories: residential, commercial, civic/amenity, and other. Additional engineered features were derived to capture internal layout and geometric complexity, including estimated internal wall counts, wall-to-wall distances \cite{ucl_space_standards18, dluhc15}, footprint shape complexity (Polsby--Popper index, see Eq.~\eqref{eq:shape_complexity}) \cite{pols1991}, and floor-to-ceiling height estimates \cite{dluhc15, simmons24}.
\begin{equation}
\label{eq:shape_complexity}
    C_b = \frac{2\sqrt{\pi A_b}}{P_b},
\end{equation}
where $C_b$ was the shape complexity (compactness) of building $b$, $A_b$ was the building footprint area, and $P_b$ was the perimeter. %Higher values of $C_b$ indicate more compact shapes, while lower values correspond to more irregular geometries.

\subsection{Indoor/Outdoor Detection and Loss Calculation}

Each \gls{UE} sample was classified as indoor or outdoor to identify O2I or I2I links. Using the bivariate Gaussian model and Eq~\eqref{eq:rel_rf_loss} defined in Section~\ref{subsec:sys_measurement}, the probability of a sample being indoor was computed by evaluating the integral of the distribution over the building polygon $\mathcal{P}_b$. A sample was classified as indoor if this overlap probability exceeded a $50\%$ threshold, or $30\%$ if the sample centroid lay within the polygon.

\subsection{Loss Labeling via Unsupervised Clustering}

To convert continuous relative loss values into categorical labels, loss distributions were first stabilized using a Box--Cox transformation. The transformed losses were then clustered using $k$-means. 

%As explained in Section~\ref{subsec:sys_problem}, this unsupervised approach avoided manually defined thresholds, which are typically environment-specific, and enabled scalable labeling in unseen regions without site-dependent calibration.

Cluster validity was assessed using silhouette scores together with sensitivity analysis over the number of clusters $k$. We evaluated $k$-means for $k \in [2,10]$. Across this range, the silhouette scores remained relatively stable at approximately $0.50$, with a marginal peak of $\approx 0.51$ at $k=3$. This indicates moderate cluster structure but limited gains in separability from increasing $k$, suggesting diminishing returns for finer partitioning. Since silhouette score alone may not fully capture practical usefulness, we further examined cluster interpretability and downstream performance. Increasing $k$ beyond 3 resulted in more fragmented clusters without clear physical meaning, while also increasing model complexity. Conversely, a two-class formulation ($k=2$), evaluated in an internal pilot study, was found to be overly coarse, merging buildings with moderate and severe attenuation into a single class.

From a deployment perspective, a small number of discrete loss categories is preferred for efficient integration into large-scale network planning and digital twin simulations, where computational efficiency and interpretability are critical. Compared to regression-based approaches, discrete classification reduces inference complexity and latency and can be suitable for practical decision-making.
Based on the combined evidence from clustering stability, interpretability, and deployment constraints, we adopted $k=3$, corresponding to low, medium, and high relative loss.

\subsection{Model Learning and Inference}

Loss labels were aggregated to the building level using z-score normalization $z_s = \frac{l_s - \mu_b}{\sigma_b}$ where $l_s$ denoted the sample-level loss, and $\mu_b$ and $\sigma_b$ represented the building-level mean and standard deviation, respectively.

\paragraph{Supervised Ensemble Training (SL)}
For buildings with sufficient measurement density, four ensemble models were trained: \gls{RanF}, \gls{XGBoost}, \gls{LightGBM}, and an integrated Voting Classifier. The models used the features listed in Table~\ref{tab:input_features} to map structural proxies to one of the three loss categories at the sample level.

The building-level label was assigned via majority voting across the sample-level predictions, ensuring consistent classification per frequency band. Fig.~\ref{fig:building_majority_vote} illustrates this process.

% -------------------------------------
\begin{table*}[t]
\normalsize	
\centering
\setlength{\tabcolsep}{9pt}
\renewcommand{\arraystretch}{1.25}

\caption{Building-Level Input Features for O2I and I2I Loss Modeling}
\label{tab:input_features}

\resizebox{\textwidth}{!}{
\begin{tabular}{
    >{\centering\arraybackslash}m{0.22\textwidth} |
    >{\centering\arraybackslash}m{0.08\textwidth} |
    >{\raggedright\arraybackslash}m{0.62\textwidth}
}
\hline
\textbf{Feature} & \textbf{O2I / I2I} & \textbf{Rationale} \\
\hline

Frequency band &
Both &
Penetration and indoor attenuation are strongly frequency dependent, with higher bands exhibiting larger losses. \\

\rowcolor[gray]{0.93}
Unit energy consumption (mean) &
Both &
Correlates with insulation quality and envelope tightness, which affect wall and internal penetration losses. \\

Unit energy consumption (std) &
Both &
Captures heterogeneity in construction quality and internal layout within a building. \\

\rowcolor[gray]{0.93}
Building usage type &
Both &
Indicative of distinct construction materials, layouts, and partition densities. \\

Wall type &
O2I &
Outer wall structure directly determines building penetration loss. \\

\rowcolor[gray]{0.93}
Wall material &
Both &
Material composition governs RF absorption and reflection at exterior and interior boundaries. \\

Wall insulation type &
O2I &
Insulation layers significantly increase O2I attenuation. \\

\rowcolor[gray]{0.93}
Window glazing type &
O2I &
Glazing technology is a dominant factor in façade penetration loss. \\

Construction year &
Both &
Newer buildings typically employ improved insulation and modern materials affecting both O2I and I2I loss. \\

\rowcolor[gray]{0.93}
Building height &
Both &
Influences diffraction, vertical propagation paths, and inter-floor attenuation. \\

Building footprint area &
Both &
Larger footprints often imply longer indoor paths and more attenuation surfaces. \\

\rowcolor[gray]{0.93}
Estimated inner wall count &
I2I &
Primary contributor to cumulative I2I loss across rooms and corridors. \\

Floor count &
I2I &
Relates to vertical propagation and floor slab penetration losses. \\

\rowcolor[gray]{0.93}
Wall-to-wall distance &
I2I &
Approximates room size and internal wall density affecting indoor signal decay. \\

Floor-to-ceiling height &
I2I &
Affects indoor propagation distance and vertical attenuation characteristics. \\

\rowcolor[gray]{0.93}
Footprint shape complexity &
Both &
Irregular geometries increase scattering and multipath effects. \\

EPC rating &
Both &
Energy-efficiency indicators, indirectly reflecting material and structural properties. \\

\hline
\end{tabular}
}
\end{table*}

% -------------------------------------

\begin{figure*}[!bt]
    \centering
    \includegraphics[width=\textwidth]{figures/majority_vote_v4_bg.png}
    \caption[Majority Vote Mechanism for Building-Level Loss Classification]{Sample-level losses $y_{s,f}$ (low, medium, high) across different serving cells $c_{s}$ and frequency bands $f$ are aggregated using public geospatial data and ensemble ML with majority voting to assign each building a final loss $y_{b,f}$ category per frequency band $f$.}
    \label{fig:building_majority_vote}
\end{figure*}

\paragraph{Semi-Supervised Learning (SSL or semi-SL)}
For buildings with sparse or missing measurements, an iterative \gls{SSL} framework was employed. High-confidence predictions from the supervised models were used as pseudo-labels and were constrained by domain-informed rules (\eg, buildings with modern Low-E glazing were weakly guided toward higher attenuation categories). This approach enabled inference across the entire study area $\mathcal{A}$, including regions with limited measurement coverage.

Although experiments were conducted within a single urban area, all building-level components of the proposed framework relied exclusively on publicly available data sources \eg, \gls{OSM}, \gls{LBSM2}) and were directly transferable to other cities with comparable datasets.

%% file: chapters/results.tex
This section presents the evaluation results of the proposed building-level O2I and I2I loss inference framework. We first describe the evaluation methodology, followed by performance analysis and comparison with baseline approaches. Feature importance, post-analysis of inferred losses, and manual validation are then used to interpret the results.

\subsection{Evaluation Methodology}

\subsubsection{Evaluation Scenarios}
We considered two inference scenarios:
\begin{itemize}
    \item \textbf{Supervised Learning (SL):} buildings with sufficient crowdsourced measurements, where loss labels were directly learned from observed data;
    \item \textbf{Semi-Supervised Learning (SSL):} buildings with sparse or missing measurements, where high-confidence predictions were iteratively incorporated as pseudo-labels.
\end{itemize}

Both scenarios were evaluated independently for \gls{O2I} and \gls{I2I} loss classification.

\subsubsection{Evaluation Metrics}
Model performance was assessed using accuracy, F1-score, prediction entropy (see Eq~\eqref{eq:eval_entropy}), and \gls{MMPP} (see Eq~\eqref{eq:eval_mmpp}). Accuracy and F1-score quantified classification performance, while entropy and MMPP characterized predictive uncertainty and confidence. The building-level entropy $\bar{H}_b$ was defined as the mean entropy across all samples in the set of samples for building $b$, $S_b$:
\begin{equation}
\label{eq:eval_entropy}
\bar{H}_b =
\frac{1}{|S_b|}
\sum_{s \in S_b}
\left(
-\sum_{c=1}^{C}
P(y_s=c|x_s)\log_2 P(y_s=c|x_s)
\right)
\end{equation}
where $C$ was the number of classes and $P(y_s=c|x_s)$ was the predicted probability for class $c$. The \gls{MMPP} was given by
\begin{equation}
\label{eq:eval_mmpp}
\mathrm{MMPP} = \frac{1}{N}\sum_{s=1}^{N} \max_{c} P(y_s=c|x_s),
\end{equation}
where $N$ denoted the number of evaluated samples. Higher MMPP and lower entropy indicated more confident and stable predictions.

\subsection{Model Training and Inference Performance}

\begin{table*}[t]
\centering
\normalsize
\caption{Performance Summary for \gls{SL} and \gls{SSL} Models}
\label{tab:perf_summary}
\renewcommand{\arraystretch}{1.25}
\setlength{\tabcolsep}{18pt}

\resizebox{\textwidth}{!}{
\begin{tabular}{l|cccc|cccc}
\hline
\multirow{2}{*}{\textbf{Model}} 
 & \multicolumn{4}{c|}{\textbf{O2I Loss}} 
 & \multicolumn{4}{c}{\textbf{I2I Loss}} \\
\cline{2-9}
 & Acc. & F1 & Entropy & MMPP 
 & Acc. & F1 & Entropy & MMPP \\
\hline

\rowcolor[gray]{0.93}
RanF (SL)       
& 0.84 & 0.80 & 1.46 & 0.48 
& 0.84 & 0.79 & 1.49 & 0.46 \\

XGBoost (SL)      
& 0.88 & \textbf{0.82} & 1.43 & 0.52
& 0.87 & 0.81 & 1.48 & 0.48 \\

\rowcolor[gray]{0.93}
LGBM (SL)     
& 0.87 & 0.81 & 1.19 & 0.63 
& 0.87 & 0.80 & 1.32 & 0.58 \\

Voting (SL)   
& 0.87 & 0.81 & 1.35 & 0.56 
& 0.87 & 0.80 & 1.43 & 0.52 \\

\rowcolor[gray]{0.93}
RanF (SSL)      
& 0.96 & 0.71 & 1.15 & 0.61
& 0.90 & 0.81 & 1.45 & 0.48 \\

XGBoost (SSL)     
& \textbf{0.98} & 0.79 & \textbf{1.09} & \textbf{0.65}
& \textbf{0.95} & \textbf{0.84} & 1.36 & 0.55 \\

\rowcolor[gray]{0.93}
LGBM (SSL)    
& 0.95 & 0.69 & 1.10 & \textbf{0.65}
& 0.90 & 0.81 & \textbf{1.28} & \textbf{0.59} \\

Voting (SSL)  
& 0.95 & 0.67 & 1.19 & 0.61
& 0.89 & 0.80 & 1.41 & 0.52 \\
\hline
\end{tabular}
}
\end{table*}

\subsubsection{Accuracy and F1-Score}

Table~\ref{tab:perf_summary} summarizes the classification performance of all \gls{SL} and \gls{SSL} models. All supervised models achieved accuracies above 0.84 for both \gls{O2I} and \gls{I2I} tasks. Semi-supervised models consistently improved accuracy, particularly for \gls{O2I}, where \textbf{SSL XGBoost} achieved the highest accuracy of 0.98.

F1-scores remained stable across methods ($\geq 0.67$). For \gls{O2I}, supervised models achieved slightly higher F1-scores, while for \gls{I2I}, \textbf{SSL XGBoost} achieved the highest F1-score (0.84). These results indicate that SSL improves overall correctness while maintaining competitive class balance.

\subsubsection{Prediction Confidence and Uncertainty}

These parameters were evaluated using MMPP and entropy. For \gls{O2I}, \textbf{SSL XGBoost} and \textbf{SSL LightGBM} achieved the highest MMPP values (0.65), with \textbf{SSL XGBoost} exhibiting the lowest entropy (1.09), indicating more confident and stable predictions. Relative to the best supervised baseline (SL LGBM), SSL XGBoost improved accuracy from 0.87 to 0.98 (12.6\% relative gain) while reducing entropy by approximately 8.4\%. %, demonstrating the benefit of SSL under limited labeled data conditions.

For \gls{I2I}, \textbf{SSL LightGBM} achieved the highest MMPP (0.59) and the lowest entropy (1.28), corresponding to a 3.4\% relative accuracy improvement and a 3.0\% entropy reduction compared to supervised-only inference. Although SSL XGBoost achieved slightly higher accuracy and F1-score, SSL LightGBM provided more confident and better-calibrated predictions (lower entropy, higher MMPP) and was therefore selected as the final model. %Overall, SSL models exhibited lower uncertainty and improved confidence.

\subsubsection{Comparison to Related Work}
%The proposed framework has used ensemble classifiers including \gls{RanF}, \gls{XGBoost}, \gls{LightGBM}, and a majority voting classifier. 
Direct numerical comparison with prior work is challenging, as existing \gls{O2I}/\gls{I2I} studies typically focus on link-level penetration loss modeling under controlled conditions, rely on manually defined attenuation thresholds, or require detailed Tx and material information. Instead, we benchmarked conceptually against representative empirical, deterministic, and learning-based approaches widely adopted in the literature, summarized in Table~\ref{tab:baseline_comparison}.

\subsection{Feature Importance}
Based on inference performance, \textbf{SSL XGBoost} and \textbf{SSL LightGBM} were selected for feature importance analysis (Table~\ref{tab:feat_import}). For \gls{O2I}, the most influential features included wall insulation type, construction year, window glazing type, and EPC rating. For \gls{I2I}, dominant features were building height, area, shape complexity, and floor-to-ceiling height.

XGBoost tended to rely on a smaller set of highly discriminative features, while LightGBM distributed importance more evenly, suggesting improved robustness when input features are noisy or incomplete.

\begin{table*}[t]
\centering
\normalsize	
\caption{O2I/I2I feature importance for SSL XGBoost and LightGBM models.}
\label{tab:feat_import}
\renewcommand{\arraystretch}{1.25}
\setlength{\tabcolsep}{18pt}

\resizebox{\textwidth}{!}{
\begin{tabular}{l|cc|cc}
\hline
\multirow{2}{*}{\textbf{Feature}} 
  & \multicolumn{2}{c|}{\textbf{O2I}} 
  & \multicolumn{2}{c}{\textbf{I2I}} \\
\cline{2-5}
 & \textbf{XGBoost (SSL)} & \textbf{LGBM (SSL)} 
 & \textbf{XGBoost (SSL)} & \textbf{LGBM (SSL)} \\
\hline
\rowcolor[gray]{0.93}
frequency band & 0.006 & \textbf{0.097} & 0.019 & 0.081 \\
unit energy consumption mean & 0.007 & 0.074 & 0.027 & 0.099 \\
\rowcolor[gray]{0.93}
unit energy consumption std & 0.006 & 0.077 & 0.024 & 0.105 \\
building usage type & 0.006 & 0.015 & 0.023 & 0.008 \\
\rowcolor[gray]{0.93}
wall type & 0.013 & 0.014 & 0.031 & 0.017 \\
construction year & \textbf{0.146} & \textbf{0.096} & \textbf{0.067} & 0.046 \\
\rowcolor[gray]{0.93}
wall insulation type & \textbf{0.656} & 0.014 & 0.025 & 0.012 \\
window glazing type & \textbf{0.051} & 0.048 & 0.028 & 0.035 \\
\rowcolor[gray]{0.93}
building height & 0.017 & \textbf{0.113} & \textbf{0.035} & \textbf{0.117} \\
building area & 0.006 & 0.093 & 0.027 & \textbf{0.114} \\
\rowcolor[gray]{0.93}
est inner walls count & 0.005 & 0.019 & \textbf{0.038} & 0.032 \\
wall material & 0.008 & 0.031 & 0.028 & 0.035 \\
\rowcolor[gray]{0.93}
floor count & 0.007 & 0.050 & 0.026 & 0.060 \\
number of vertices in footprint & 0.007 & 0.088 & --- & --- \\
\rowcolor[gray]{0.93}
shape complexity in footprint & 0.005 & \textbf{0.100} & 0.025 & \textbf{0.133} \\
wall-to-wall distance & --- & --- & 0.026 & 0.012 \\
\rowcolor[gray]{0.93}
floor-to-ceiling height & --- & --- & 0.024 & \textbf{0.108} \\
EPC rating & \textbf{0.054} & 0.072 & \textbf{0.580} & 0.034 \\
\hline
\end{tabular}
}
\end{table*}

\subsection{Post-Analysis on Predicted Loss}
Based on the feature importance results, a post-analysis was performed to explore how key building characteristics were associated with the predicted building-level O2I and I2I losses. For O2I loss, the analysis focused on \textit{construction year}, \textit{EPC rating}, \textit{wall insulation type}, and \textit{window glazing type}. For I2I loss, the main features examined were \textit{building height}, \textit{building area}, \textit{shape complexity in footprint}, and \textit{est. inner walls count}. The observed relationships are discussed below using the figures referenced in each subsection.

\subsubsection{O2I Loss (SSL XGBoost)}
\begin{figure}[!ht]
    \centering
    \begin{subfigure}[b]{\columnwidth}
        \centering
        \includegraphics[width=\linewidth]{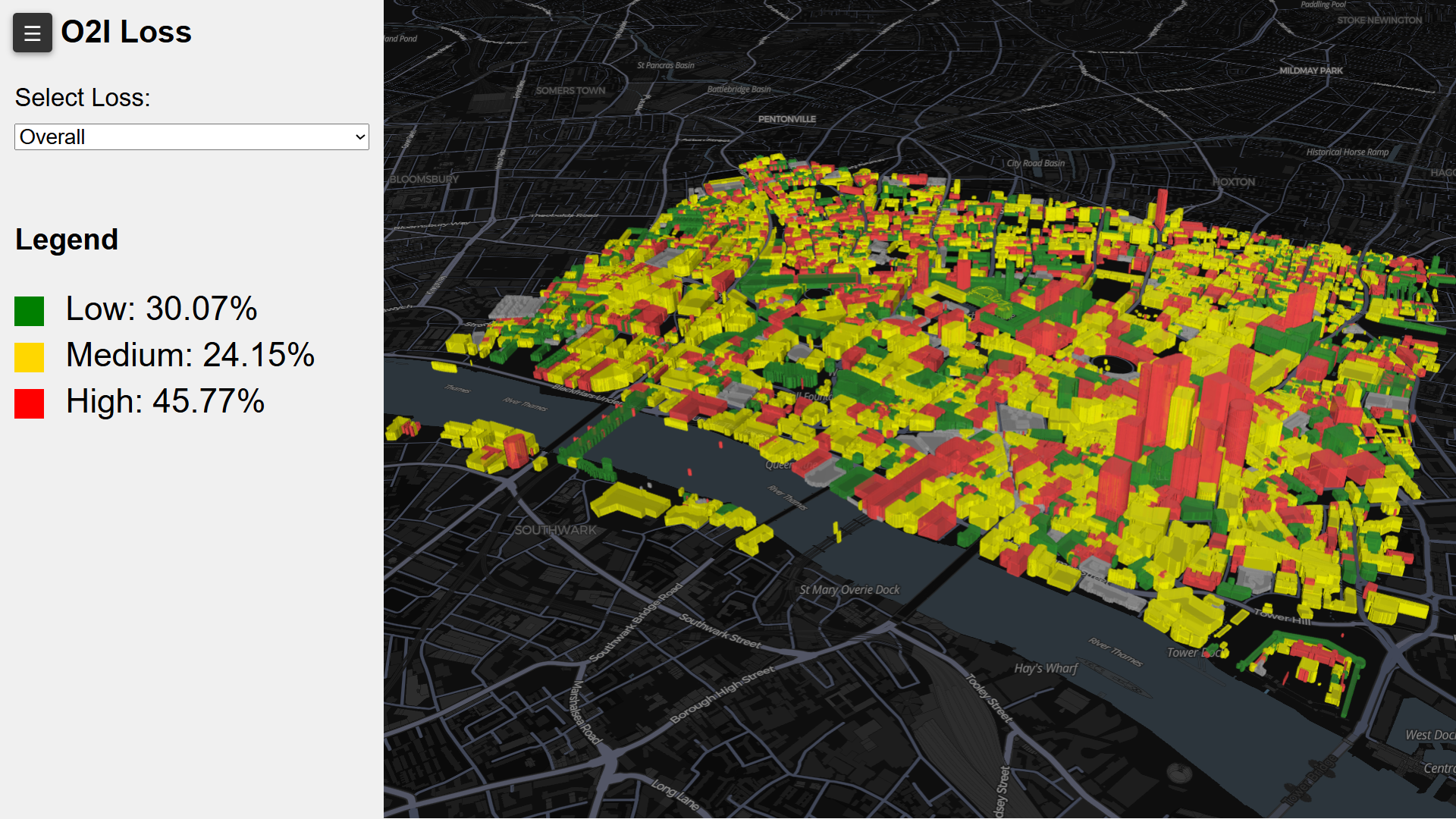}
    \caption[O2I Loss Distribution on Map]{O2I loss}
        \label{fig:o2i_loss_map_overall}
    \end{subfigure}
    \hfill
    \begin{subfigure}[b]{\columnwidth}
        \centering
        \includegraphics[width=\linewidth]{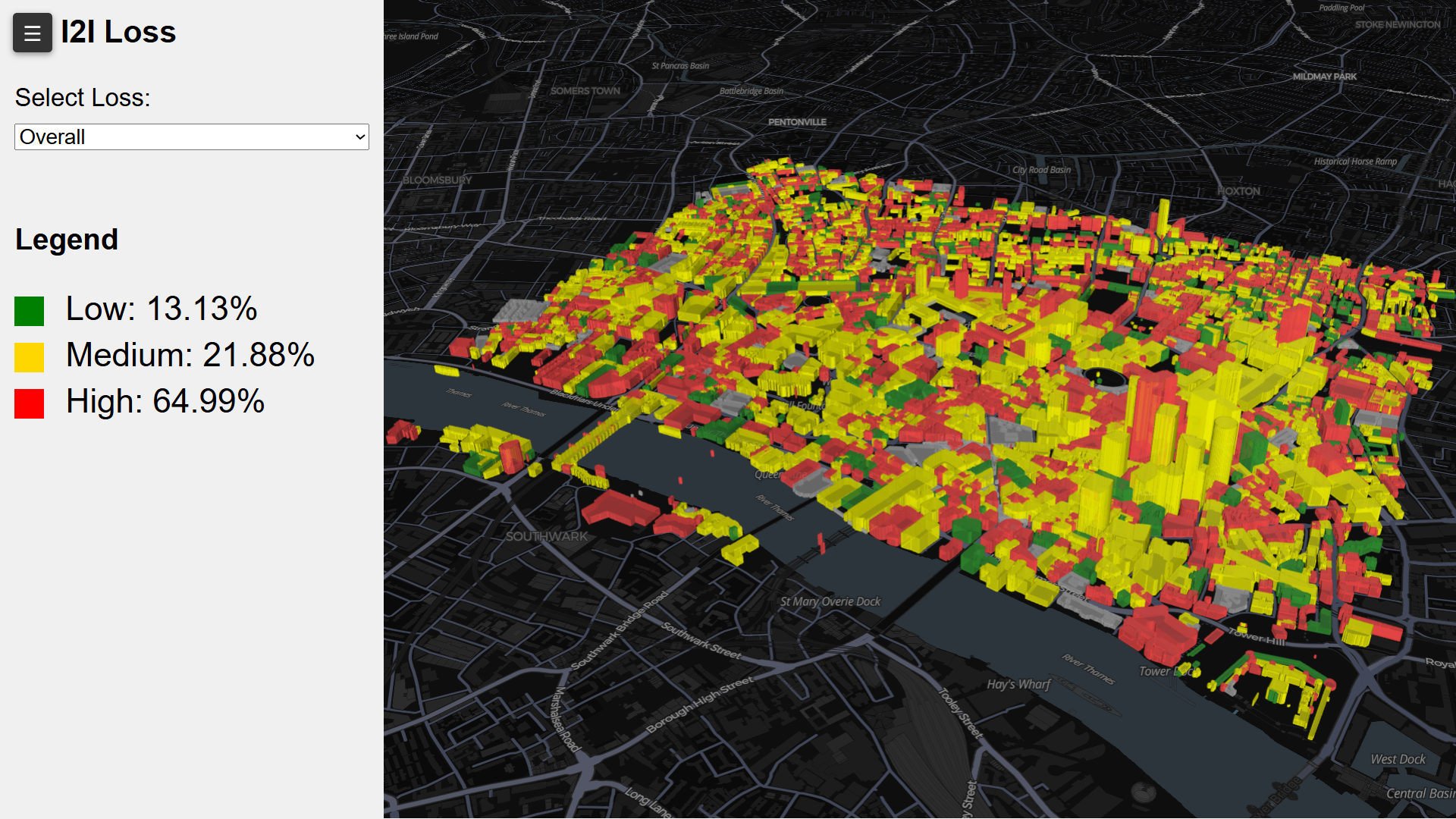}
    \caption[I2I Loss Distribution on Map]{I2I loss}
        \label{fig:i2i_loss_map_overall}
    \end{subfigure}
	\caption{O2I and I2I loss distribution of the test region.}
    \label{fig:o2i_i2i_loss_map_overall}
\end{figure}

\begin{figure}[!ht]
    \centering
    \begin{subfigure}[b]{\columnwidth}
        \centering
        \includegraphics[width=\linewidth]{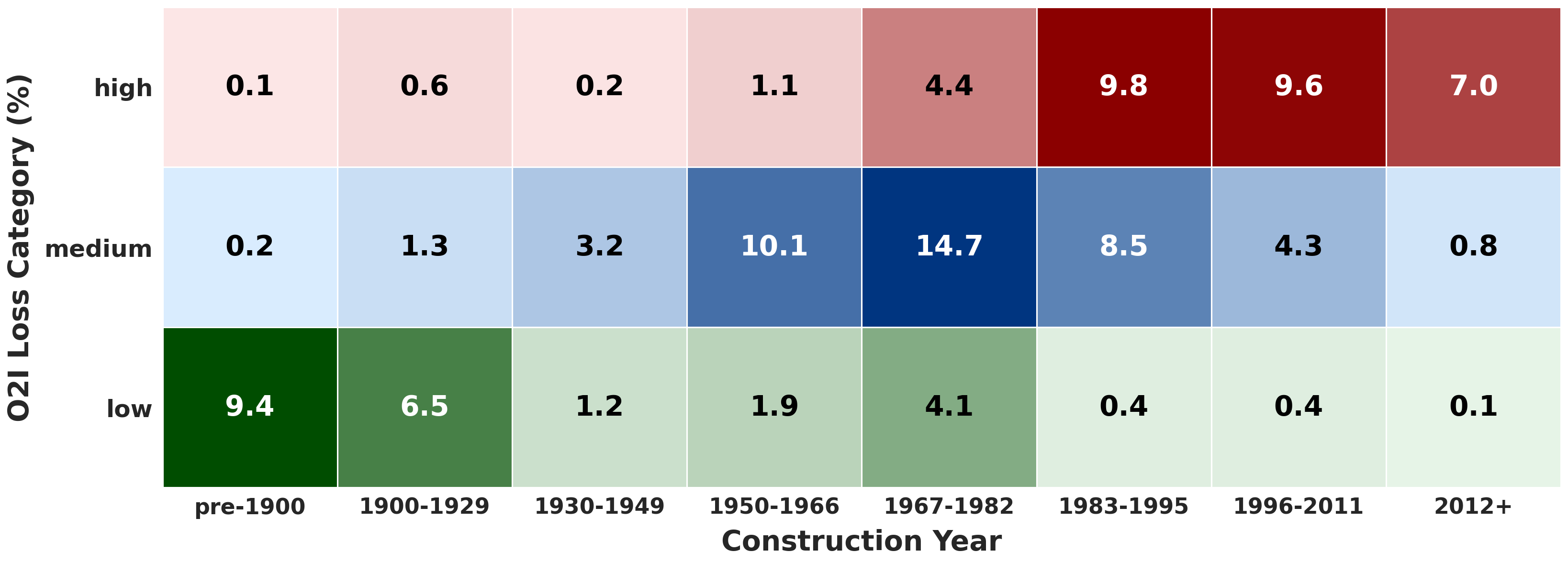}
        \caption{Construction Period}
        \label{fig:o2i_year_heatmap}
    \end{subfigure}
    \hfill
    \begin{subfigure}[b]{\columnwidth}
        \centering
        \includegraphics[width=\linewidth]{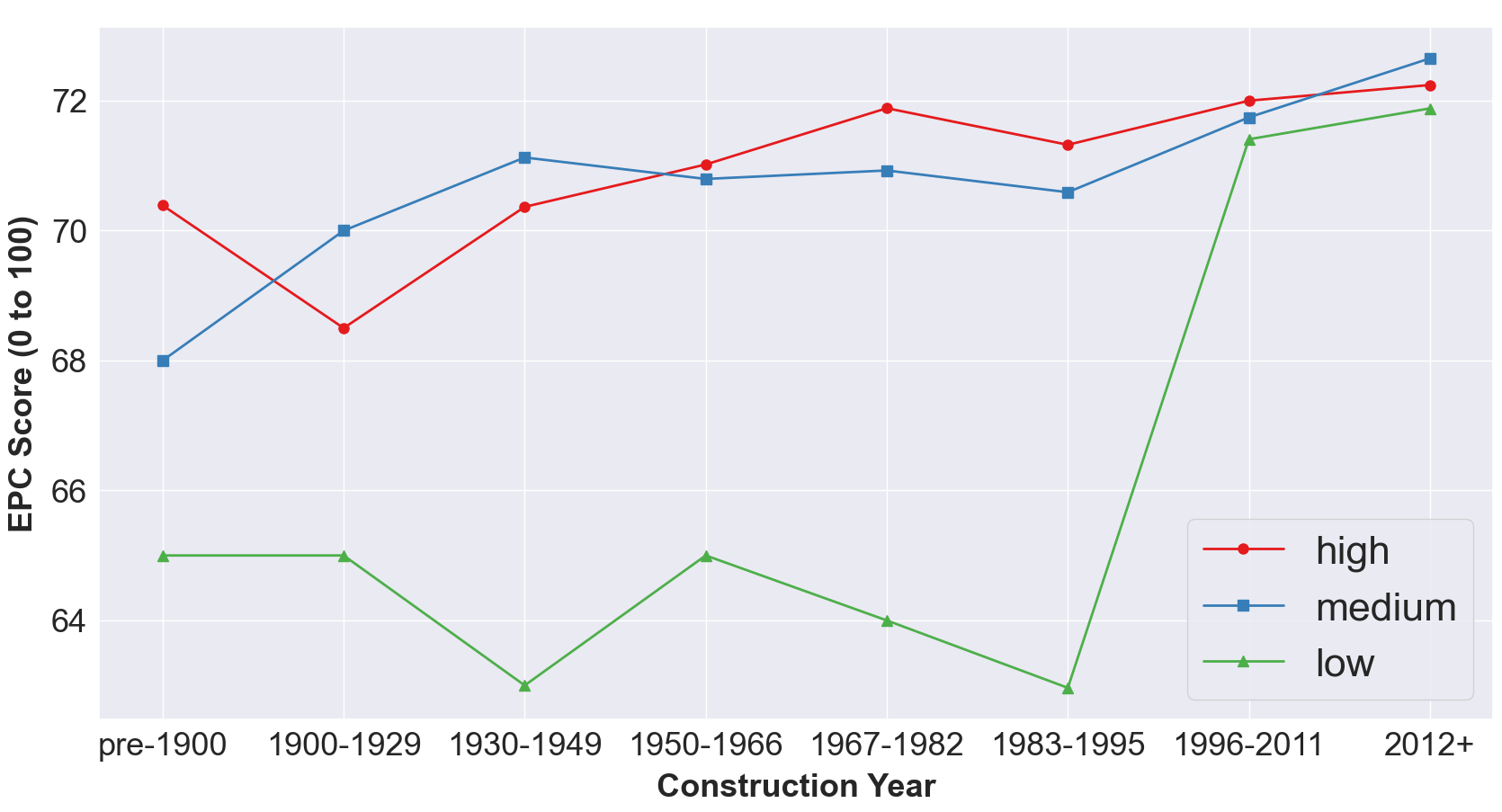}
        \caption{EPC Scores}
        \label{fig:o2i_epc_line}
    \end{subfigure}
    \hfill
    \begin{subfigure}[b]{\columnwidth}
        \centering
        \includegraphics[width=\linewidth]{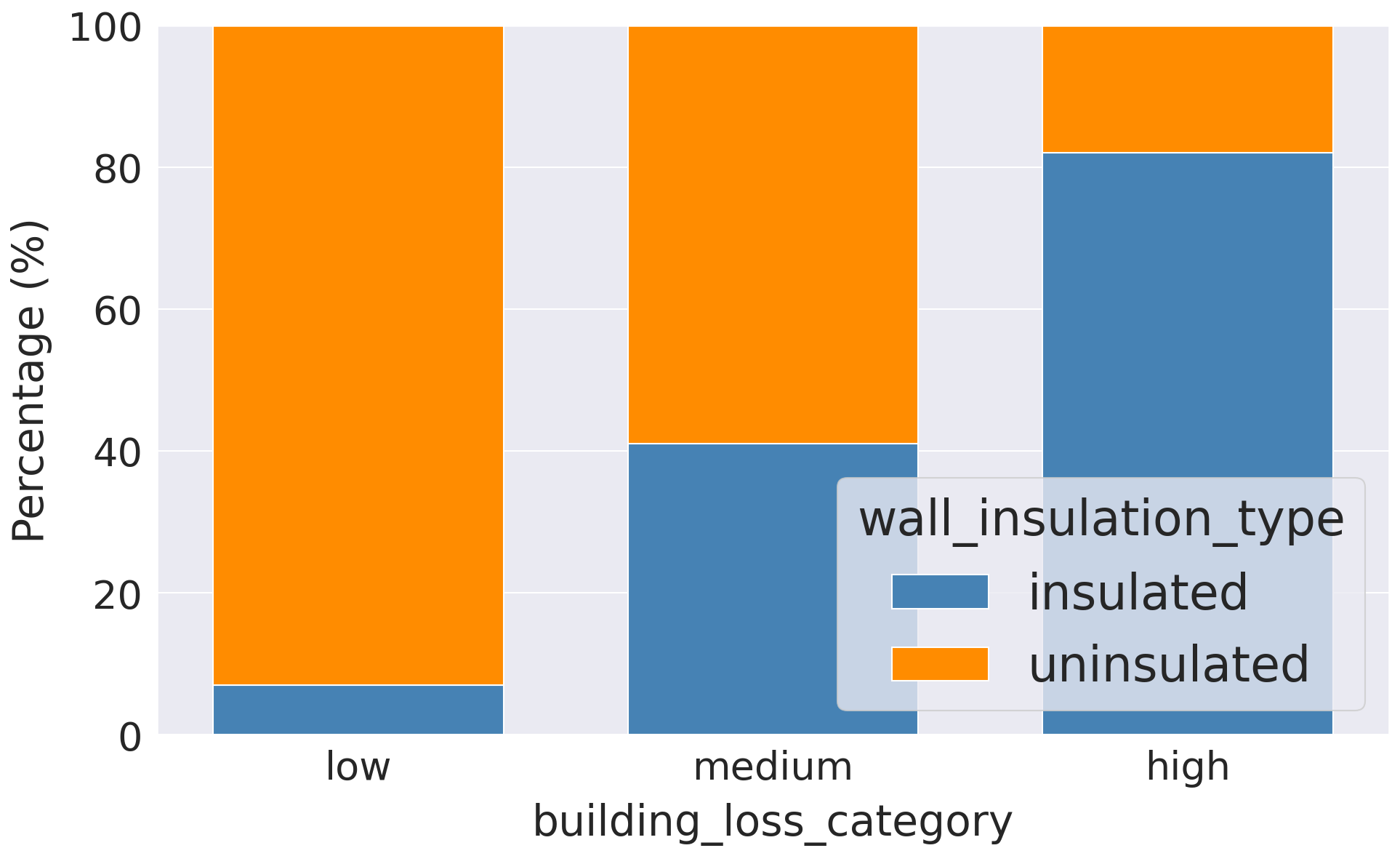}
        \caption{Wall Insulation}
        \label{fig:o2i_insulation_stacked_bar}
    \end{subfigure}
    \hfill
    \begin{subfigure}[b]{\columnwidth}
        \centering
        \includegraphics[width=\linewidth]{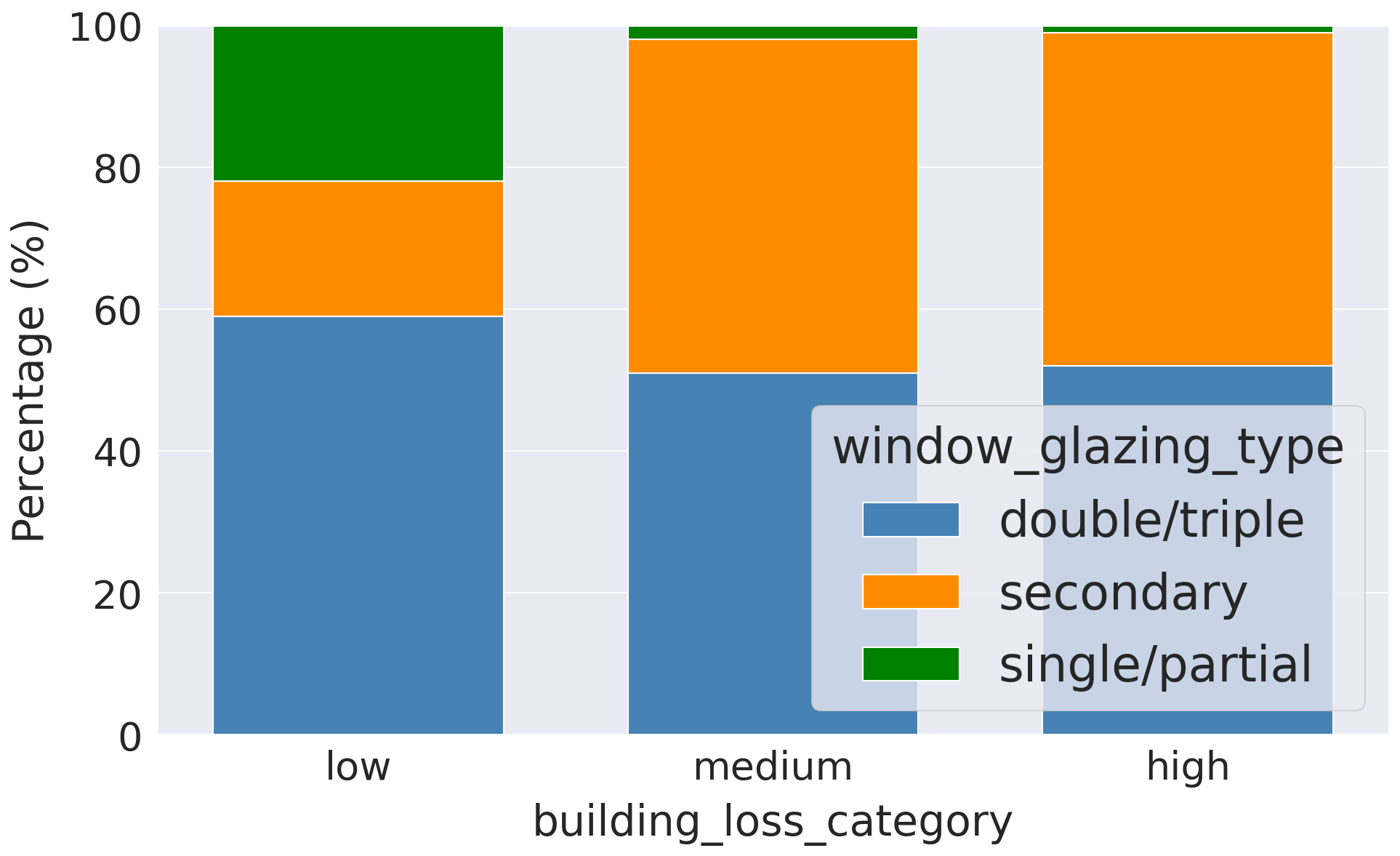}
        \caption{Window Glazing Type}
        \label{fig:o2i_glazing_stacked_bar}
    \end{subfigure}
    \caption{O2I Post-Analysis (SSL XGBoost)}
    \label{fig:o2i_post_analysis}
\end{figure}

The inferred O2I loss distribution consisted of high (45.77\%), medium (24.15\%), and low (30.07\%), as illustrated in Fig.~\ref{fig:o2i_loss_map_overall}. 
Clear trends were observed between O2I loss and building characteristics. High loss values were primarily associated with newer buildings (Fig.~\ref{fig:o2i_year_heatmap}) exhibiting high EPC ratings (Fig.~\ref{fig:o2i_epc_line}), insulated walls (Fig.~\ref{fig:o2i_insulation_stacked_bar}), and modern multi-layer glazing (Fig.~\ref{fig:o2i_glazing_stacked_bar}). These features, which are commonly associated with energy-efficient design, appeared to increase the attenuation of RF signals entering the building. In contrast, older buildings with limited or no wall insulation, single-glazed windows, and lower EPC ratings generally exhibited lower O2I loss.
When analyzed by EPC rating (Fig.~\ref{fig:o2i_epc_line}), it was observed that for buildings constructed after 1996, the EPC scores of low-loss buildings increased sharply, although they remained lower than those of medium- and high-loss buildings. This increase likely reflected stricter energy-efficiency regulations and widespread retrofitting in newer buildings, which improved thermal performance even in structures with relatively low O2I attenuation.
Overall, the results suggested that improvements in energy efficiency were closely linked to increased O2I signal attenuation.

%------------------------------------

% ------------------------------------------

\subsubsection{I2I Loss (SSL LightGBM)}
The inferred I2I distribution was high (64.99\%), medium (21.88\%), and low (13.13\%), illustrated in Fig.~\ref{fig:i2i_loss_map_overall}. 

\begin{figure}[!ht]
    \centering
    % First row: two subfigures side by side
    \begin{subfigure}[b]{\columnwidth}
        \centering
        \includegraphics[width=\linewidth]{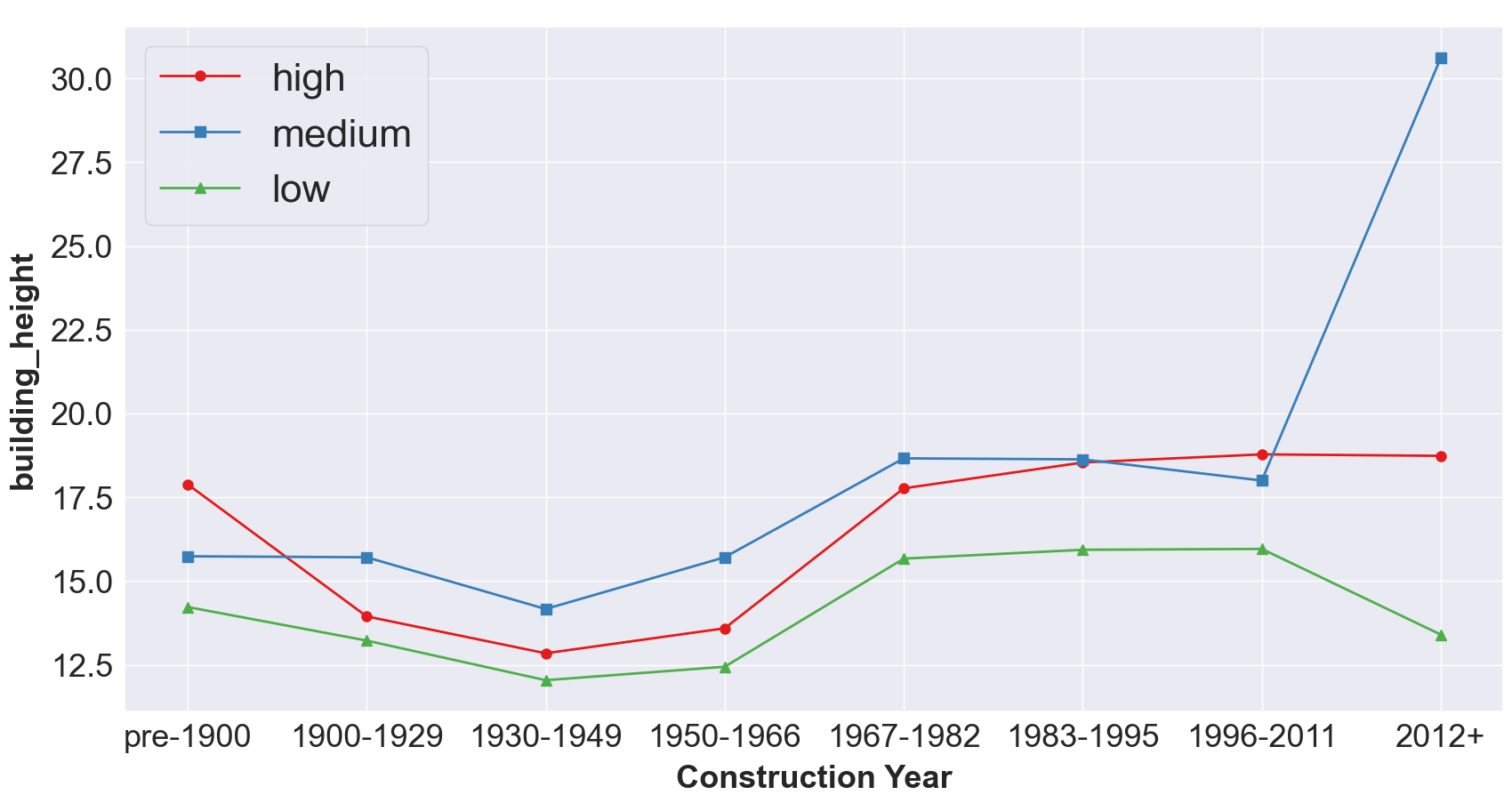}
        \caption{Building Height}
        \label{fig:i2i_height_line}
    \end{subfigure}
    %\vspace{0.5em}
    \hfill
    \begin{subfigure}[b]{\columnwidth}
        \centering
        \includegraphics[width=\linewidth]{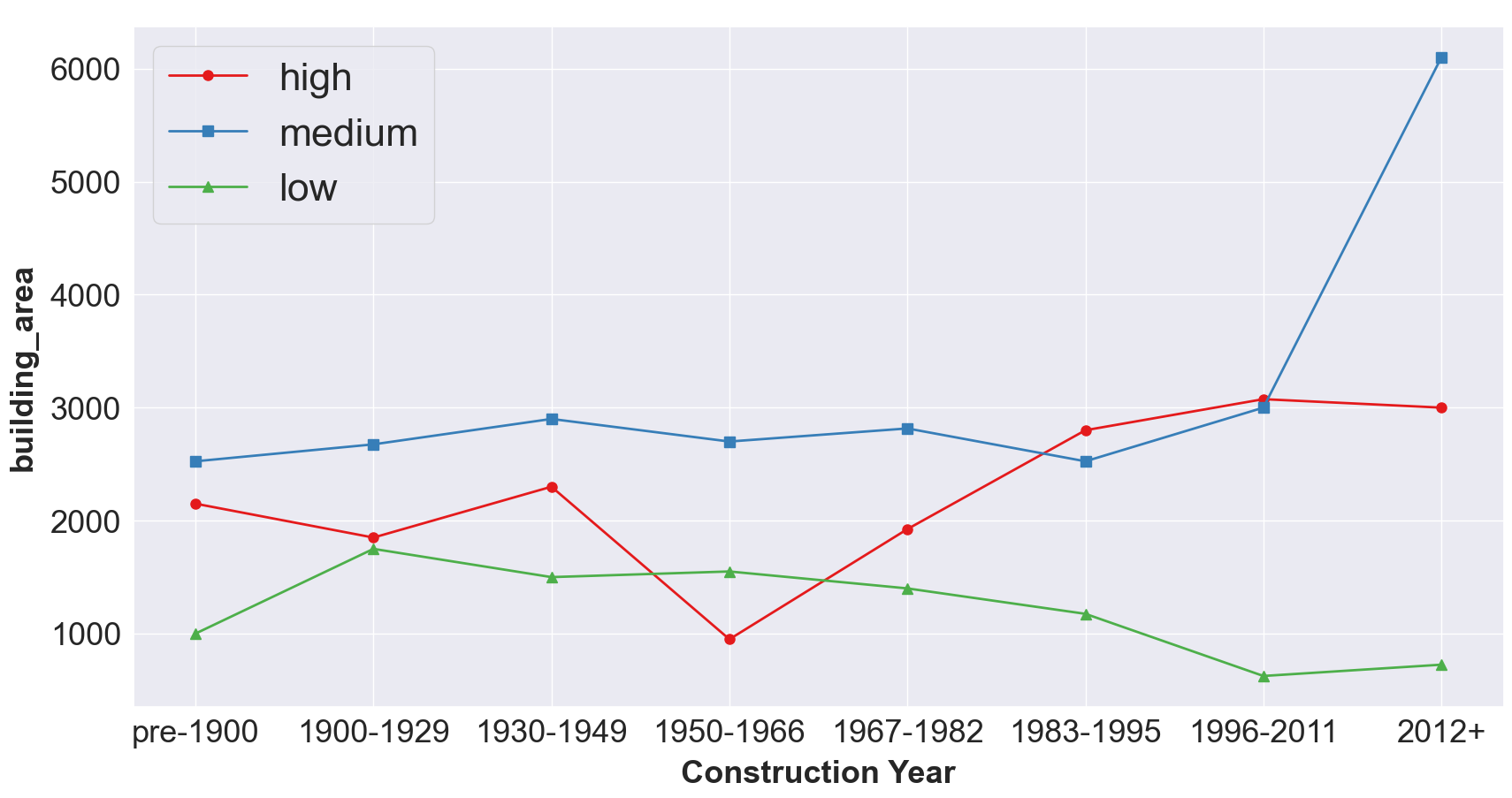}
        \caption{Building Area}
        \label{fig:i2i_area_line}
    \end{subfigure}
    % Vertical spacing
    %\vspace{0.5em}
    % Second row: third subfigure (full width of column)
    \begin{subfigure}[b]{\columnwidth}
        \centering
        \includegraphics[width=\linewidth]{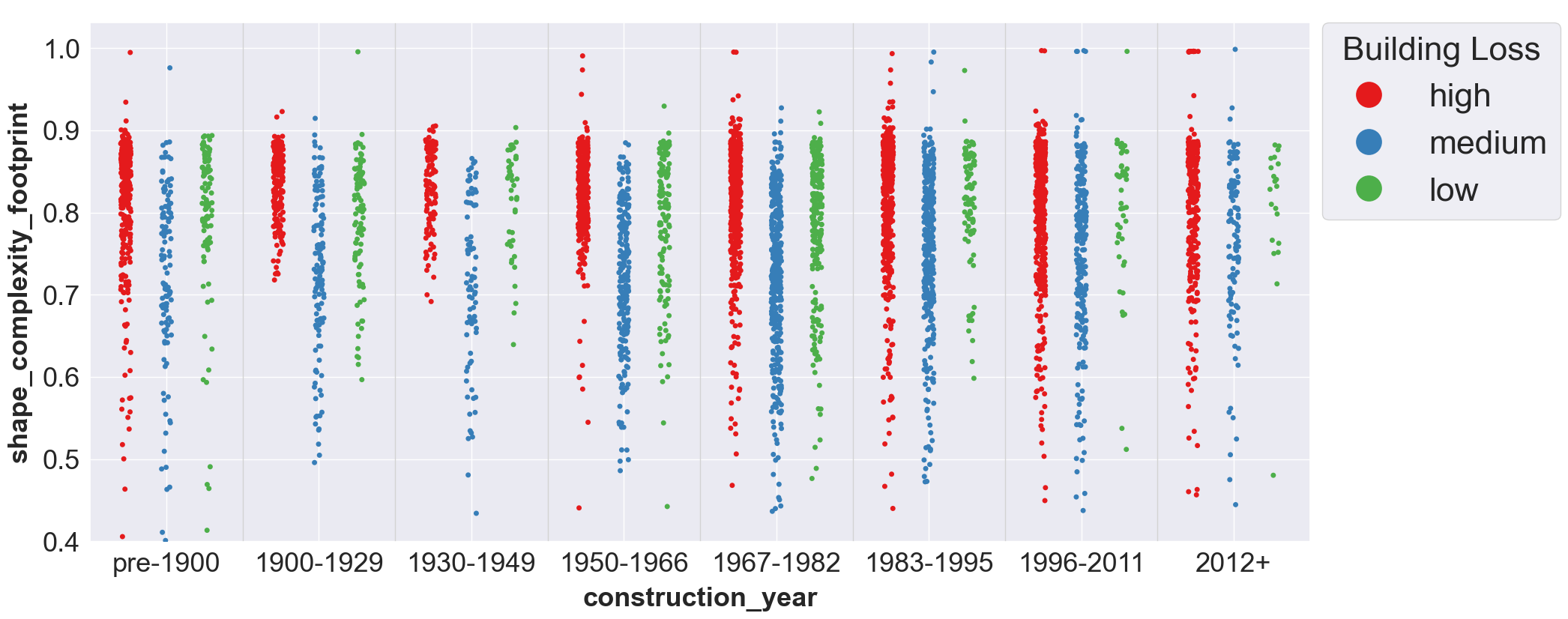}
        \caption{Shape Complexity}
        \label{fig:i2i_shape_complexity_scatter}
    \end{subfigure}
    % Vertical spacing
    %\vspace{1em}
    % Third row: fourth subfigure (full width of column)
    \begin{subfigure}[b]{\columnwidth}
        \centering
        \includegraphics[width=\linewidth]{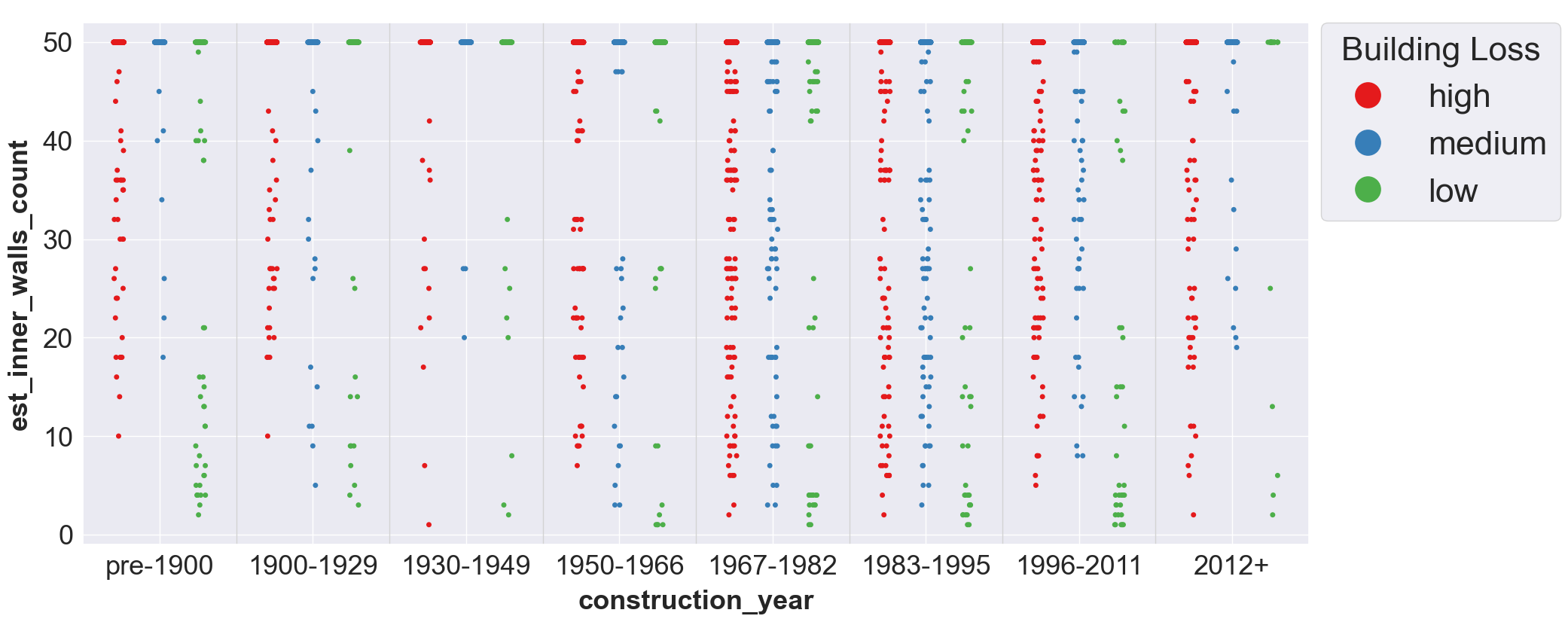}
        \caption{Est. Inner Walls Count}
        \label{fig:i2i_est_inner_wall_count_scatter}
    \end{subfigure}
    % Main caption
    \caption[I2I Post-Analysis, SSL LightGBM]{I2I Post-Analysis (SSL LightGBM).}
    \label{fig:i2i_post_analysis}
\end{figure}

Medium and high I2I loss buildings tended to be taller (Fig.~\ref{fig:i2i_height_line}) and larger (Fig.~\ref{fig:i2i_area_line}) across construction periods, with a notable increase in height and area for medium-loss buildings constructed after 2012, indicating that building height and area were important factors for I2I attenuation, particularly in modern structures.

Footprint shape complexity (see Fig.~\ref{fig:i2i_shape_complexity_scatter}) showed no clear trend across all loss categories; however, within the higher complexity range (0.9–1.0), medium and high-loss buildings were more common than low-loss buildings. Overall, I2I loss appeared to be influenced more by construction year than by footprint complexity, as newer buildings were more likely to exhibit medium or high loss.

In older buildings, higher I2I loss was associated with a greater estimated number of inner walls (see Fig.~\ref{fig:i2i_est_inner_wall_count_scatter}), highlighting the role of internal partitions. For newer (post-1966) buildings, high I2I loss occurred regardless of inner-wall count, likely due to the combined effects of taller structures, complex layouts, and dense construction materials. Since detailed inner-wall material data were unavailable, these interpretations were based on typical architectural practices and observed trends.

%------------------------------------

\subsubsection{Building Usage, Frequency Band, and Wall Material}
No clear correlation was observed between building usage, frequency band, or wall material and O2I/I2I loss, likely due to sparse, noisy, and imbalanced crowdsourced data. These factors complicated MICE imputation and may lower model prediction accuracy.
%No explicit correlations of building usage, frequency band, and wall material were found with O2I/I2I loss. This lack of correlation is likely due to the inherent data limitations: imbalanced frequency band distribution and sparse/noisy material and usage information sourced from the crowdsourced OSM dataset. These factors complicated MICE imputation and may lower model prediction accuracy.

\subsection{Manual Validation}
Manual spot-checking was performed to qualitatively validate inferred loss categories. For example, \textit{100~Bishopsgate}, a modern glass-dominant building, was correctly classified as high \gls{O2I} loss. Misclassifications (\eg, \textit{22~Bishopsgate}) were primarily attributable to imputation uncertainty and voting tie-breaking effects, rather than systematic model bias.

%% file: chapters/conclusions.tex
This study presented a building-level O2I and I2I signal loss classification framework using passive crowdsourced UE measurements and open geospatial data with combined supervised (SL) and semi-supervised learning (SSL). The proposed SL+SSL approach improved classification accuracy by up to \textbf{12.6\% for O2I} and \textbf{3.4\% for I2I}, while consistently reducing prediction uncertainty. \textbf{SSL XGBoost} achieved the best performance for O2I classification, whereas \textbf{SSL LightGBM} performed best for I2I loss. The result shows that O2I loss is strongly linked to building age and energy-efficiency features, while I2I loss is mainly influenced by building height, area, construction year, and internal structural characteristics, highlighting the critical role of building properties in signal attenuation. These findings demonstrate that integrating open geospatial data, passive measurements, and SSL models enables scalable building-level indoor coverage analysis, offering practical insights for wireless network planning. From a deployment perspective, the adopted discrete classification scheme and gradient boosting models offer favorable inference latency and low computational overhead compared to regression-based alternatives, making the framework well-suited for integration with existing network planning pipelines using openly available geospatial data. This supports near–real-time or large-scale offline analysis without additional measurement campaigns. Future work can incorporate richer indoor building data, extend analysis to higher-frequency bands, and validate the approach across additional regions.

%Key limitations include the lack of detailed internal layout and wall material data, limited measurements at higher frequency bands, and reliance on passive crowdsourced data without on-site validation. 